\documentclass{article}

\usepackage{PRIMEarxiv}

\usepackage[utf8]{inputenc} 
\usepackage[T1]{fontenc}    
\usepackage{hyperref}       
\usepackage{url}            
\usepackage{booktabs}       
\usepackage{amsfonts}       
\usepackage{nicefrac}       
\usepackage{microtype}      
\usepackage{lipsum}
\usepackage{fancyhdr}       
\usepackage{graphicx}       
\graphicspath{{media/}}     
\usepackage{tabularx,tabulary}

\usepackage[algo2e,ruled,vlined]{algorithm2e} 
\usepackage{algorithm}
\usepackage{svg}

\usepackage{tabularx}

\usepackage{kbordermatrix}

\usepackage{graphicx}

\usepackage{times}
\usepackage{svg}
\usepackage{multirow}
\usepackage{fancyhdr,graphicx,amssymb}
\usepackage{subcaption}
\usepackage{float}
\usepackage{mathtools}
\DeclarePairedDelimiter\abs{\lvert}{\rvert}%

\usepackage{xtab,caption}
\usepackage{systeme}          
\usepackage[graphicx]{realboxes}
\usepackage{pdfpages}

\newcommand{\card}[1]{\ensuremath{\lvert #1\rvert}\xspace}

\newcommand{\svm}{${\tt RankingSVM}$}
\newcommand{\ahp}{${\tt AHPRank}$}
\newcommand{\svmp}{${\tt RankingSVM.0}$}
\newcommand{\ahpp}{${\tt AHPRank.0}$}
\newcommand{\svma}{${\tt RankingSVM.1}$}
\newcommand{\ahpa}{${\tt AHPRank.1}$}
\newcommand{\yules}{${\tt YulesY}$}
\newcommand{\cosine}{${\tt Cosine}$}
\newcommand{\lamb}{${\tt Lambda}$}
\newcommand{\lev}{${\tt Leverage}$}
\newcommand{\cert}{${\tt Certainty}$}
\newcommand{\lap}{${\tt Laplace}$}
\newcommand{\IF}{${\tt InterstFactor}$}

\newcommand{\nbTrans}{{$\abs{\mathcal{T}}$}}
\newcommand{\nbItems}{{$\abs{\mathcal{I}}$}}
\newcommand{\nbRules}{${\tt \#Rules}$}

\newtheorem{theorem}{Theorem}
\newtheorem{proposition}[theorem]{Proposition}%

\newtheorem{example}{Example}%

\title{Boosting the Learning for Ranking Patterns}

\author{
  Nassim Belmecheri \\
  Lab. LITIO \\
  University of Oran1 \\
  Oran, Algeria \\
  \texttt{belmecheri.nassim@edu.univ-oran1.dz} \\
   \And
  Noureddine Aribi \\
  Lab. LITIO \\
  University of Oran1 \\
  Oran, Algeria \\
  \texttt{aribi.noureddine@gmail.com} \\
  \And
  Nadjib Lazaar \\
  LIRMM \\
  University of Montpellier \\
  Montpellier, France\\
  \texttt{Nadjib.Lazaar@lirmm.fr} \\
  \And
  Yahia Lebbah \\
  Lab. LITIO \\
  University of Oran1 \\
  Oran, Algeria \\
  \texttt{ylebbah@gmail.com} \\
  \And
  Samir Loudni \\
  TASC (LS2N-CNRS)\\
  IMT Atlantique \\
  Nantes, France\\
  \texttt{samir.loudni@imt-atlantique.fr} \\
}

\begin{document}
\maketitle

\begin{abstract}
Discovering relevant patterns for a particular user remains a challenging tasks in data mining. 
Several approaches have been proposed to learn user-specific pattern ranking functions.
These approaches generalize well, but at the expense of the running time. 
On the other hand, several measures are often used to evaluate the interestingness of patterns, with the hope to reveal a  ranking that is as close as possible to the user-specific ranking. 
In this paper, we formulate the problem of learning pattern ranking functions as a multicriteria decision making problem. 
Our approach aggregates different interestingness measures into a single weighted linear ranking function, using an interactive learning procedure that operates in either passive or active modes. 
A fast learning step is used for eliciting the weights of all the measures by mean of pairwise comparisons. 

This approach is based on Analytic Hierarchy Process (AHP), and a set of user-ranked patterns to build a preference matrix, which compares the importance of measures according to the user-specific interestingness. 
A sensitivity based heuristic is proposed for the active learning mode,  
in order to insure high quality results with few user ranking queries. 
Experiments conducted on well-known datasets show that our approach significantly reduces the running time and returns precise pattern ranking, while being robust to user-error compared with state-of-the-art approaches.  

\end{abstract}


\section{Introduction}

Data mining is the study of how to extract relevant information from data and exploit it as useful knowledge. 
One of its most important subfields, pattern mining, involves searching and
enumerating interesting patterns in data.
In the last decade, the pattern mining community
has witnessed a sharp shift from efficiency-based approaches to methods that can extract more meaningful patterns.
Unfortunately, obtaining interesting results with traditional pattern mining methods can be a tough and time-consuming task. The two main issues are
that: 1) humongous amounts of patterns are found, of which many are redundant,
and 2) preferences and background knowledge of the domain expert are not taken into account.
The importance of taking user preferences and knowledge into account was first emphasized by Tuzhilin~\cite{SilberschatzT95}. 
The main idea is to model the user's preferences using objective interestingness measures. 
Here, the interestingness of a pattern is computed from the data only. 
However, as stated in \cite{Bie11}, using objective quality measures are of limited practical use, since interestingness depends on the specific user and task at hand. 

In recent years, pattern mining has been challenged by an increasing focus on {\it user-centered,
interactive, and anytime} pattern mining \cite{dzyuba2013interactive,BoleyMKTW13,dzyuba2017learning}. 
This new paradigm stresses that users should be presented quickly with patterns likely to be interesting to them, and typically affect later iterations of the interactive mining process by giving feedback. 
An important aspect of this framework is the ability to learn user-specific pattern ranking functions from feedback. 
This idea was first investigated in \cite{xin2006discovering} and recently extended by Boley et al. \cite{BoleyMKTW13} and Dzyuba et al. \cite{dzyuba2013interactive,dzyuba2017learning} in the context of interactive pattern mining. 
These approaches exploit standard machine learning techniques to learn weighted vectors according to some selected features on patterns (i.e. items, transactions, length $\ldots$). 

In \cite{dzyuba2013interactive}, a linear ranking function is learned using \svm. 
In \cite{BoleyMKTW13,dzyuba2017learning}, the authors propose to use Stochastic Coordinate Descent (SCD) \cite{Shalev-ShwartzT11} 
to learn a logistic function. 
While these methods allow to exploit user feedback to learn ranking functions, they also result in a more computationally expensive learning task, particularly when the number of pairs of patterns and measures used for ranking increases. 
However, a vital ingredient of this framework is the ability to quickly present a set of patterns to the user and focus on what should be of interest using some interestingness measures. 

In this paper, we tackle the problem of learning pattern ranking functions as a multicriteria decision making problem. 
We use a {\it weighted linear function} to aggregate all the individual measures into a global ranking function.  
A fast and scalable learning algorithm is used to maintain the right expected weights of measures. 
The proposed approach exploits the Analytical Hierarchy Process (AHP)~\cite{saaty1988analytic}, and a set of user-ranked patterns to learn the measures' weighting vector, that maximizes the correlation between the (unknown) user's ranking function and the learned AHP-based ranking function. 

This approach was firstly introduced in our earlier work \cite{RNTI/papers/1002739}, where we show initial results of the passive version of the algorithm. More precisely, our new contributions compared with the previous version are as follows:
\begin{itemize}
\item A detailed presentation of the whole approach with running examples.
\item A new generic version of the learning algorithm to learn both in active and passive modes.
\item Proposition of a sensitivity based heuristic for selecting patterns when learning in the active mode.
\item Implementing three different user simulations.
\item Validation of the whole approach with its various settings (i.e., passive and active modes, user simulations, etc.) on large and new datasets.
\item Showing the robustness of the approach in an interactive learning process, considering user mistakes and user changing its ranking criterion.
\end{itemize} 
To assess the interest of our approach, we consider the association rules mining problem as a case study. 
Experiments conducted on a wide range datasets show that our approach 
significantly reduces the running time, while ensuring accurate pattern ranking compared to the state-of-the-art approaches.

The rest of the paper is organized as follows: 
Section~2 introduces the related work.
Section~3 presents some necessary preliminaries. 
Section~4 formulates the tackled problem and describes our approach. 
Section~5 illustrates our approach on a running example. 
Empirical results are reported and discussed in Section~6.
Section~7 concludes the paper.

\section{Related work}
Recent approaches for iterative/user-centric pattern mining are based on learning ranking functions throughout user pairwise pattern rankings. This problem is known as {\it object ranking}. Several methods have been developed for the object ranking  task ~\cite{KamishimaKA10}. A common solving technique involves minimizing pairwise loss, e.g., the number of discordant pairs. 
In \cite{xin2006discovering}, the authors propose a log-linear model for itemsets to learn the user’s prior knowledge from her 
feedback. 
They use \svm{} formulations~\cite{ranking} to learn a ranking function. For more complex patterns (e.g., sequential patterns), a belief model is proposed, which exploits belief probabilities assigned to transactions.  
Later, Dzyuba et al.~\cite{vlad} extend the approach based on \svm{} for active preference learning for ranking patterns. 
According to \cite{bottou2007support}, the worst-case time complexity to solve the SVM problem varies between $k^2$ and $k^3$ on the number of samples $k$; whereas our approach is linear on $k$  (see section \ref{sec:complexity}).
In \cite{BoleyMKTW13}, the same approach is adopted to learn a logistic function using  Stochastic Coordinate Descent (SCD) \cite{Shalev-ShwartzT11}. The same technique is also used in \cite{dzyuba2017learning}, but the learned function is exploited for patterns sampling.
In \cite{bhuiyan2016priime}, regression techniques are adopted to learn ranking functions. 

Our approach is different and tackles the problem of learning pattern ranking functions as a multicriteria decision making problem using AHP. Moreover, our method remains fast when the number of patterns used for ranking increases.

\newcommand{\attrs}{\mathcal{A}}
\newcommand{\objs}{\mathcal{O}}
\newcommand{\rels}{\mathcal{L}}
\newcommand{\patts}{\mathcal{L}}
\newcommand{\ranks}{\mathcal{K}}
\def\N{\mbox{$\mathbb{N}$}}
\newcommand{\setRp}{{\mathord{{\mathbb R_{+}}}}}

\section{Background}

\subsection{Pattern mining and interestingness measures}
Data can be abstracted as a triplet ($ \objs,\attrs, \rels$) 
in which  $\objs$ is a set of objects, $\attrs$ is a set of attributes and $\rels$ is the language used to express objects on attributes. This framework is inspired from inductive databases concepts \cite{DBLP:journals/sigkdd/Raedt02,DBLP:journals/cacm/ImielinskiM96}.
Following the language $\rels$, for instance, we can get classical datasets: (1)~when $\rels$ is a binary relation between attributes and objects (i.e., $\rels \subseteq \objs\times\attrs$), we get the classical itemset mining framework; (2)~when $\rels$ expresses each object as a sequence of attributes, we get a sequential dataset; (3)~when $\rels$ expresses each object as a graph where nodes and vertices are labeled, we get a dataset of graphs.
Data mining tasks aim at discovering interesting regularities, namely patterns, between objects in large-scale datasets.

Measuring the interestingness of discovered patterns w.r.t. the user-specific preferences is an active field in Data Mining.
Nine concepts have been introduced to determine whether a pattern is interesting or not:
\emph{Conciseness, Generality, Reliability, Peculiarity, Diversity, Novelty, Surprisingness, Utility} and  \emph{Applicability} \cite{GengH06}.
Some concepts are correlated like \emph{Surprisingness} and \emph{Applicability} concepts,
 are conflicting like \emph{Generality} and \emph{Pecularity}, or independent like \emph{Novelty} and \emph{Utility}. 
The user has in mind some of these concepts to say that she prefers a pattern than another.
A wide range of measures, and aggregation of these measures, have been proposed to satisfy the user preferences. 

\subsection{ Analytical Hierarchy Process (AHP)} 
\label{sec:ahp:background}
AHP is a well-known multicriteria decision-making method developed by Saaty~\cite{saaty1988analytic}. It is based on structuring and synthesizing the decision problem into a problem hierarchy, and eliciting criteria weights from the decision maker through a series of pairwise comparisons, as opposed to utilizing numerical values directly.
After constructing a hierarchy of the decision problem, the importance weights of criteria can be calculated according to the following three main steps: 
\begin{enumerate}
	\item A user is asked to compare the criteria of a given level in a pairwise manner to estimate their relative importance in relation to criteria at the preceding level. 
	The pairwise comparisons are collected into a pairwise comparison matrix, {\bf A} = $(a_{ij})_{m\times m}$, 
	with $a_{ij}$ expressing the relative importance of criterion $i$ to criterion $j$: the 
	$i^{th}$  criterion is better than the $j^{th}$ criterion if $a_{ij}>1$.
	The preference matrix {\bf A} is filled s.t. $a_{ij}= 1/a_{ji}$ and $a_{ii}=1$. 
	For simplicity and easiness, AHP uses $9$ degrees of preference, where $a_{ij}= 1$ indicates an indifference and $a_{ij}=9$ an absolute preference.
	
	
	\item  Once the pairwise comparison matrix at a given level is built, the weight vector $w = (w_1, \ldots, w_m)^T$ can be computed using several mathematical techniques proposed in the literature. 
	For example, the eigen vector method (EVM) or distance-based minimization methods, such as Least Squares Method, Logarithmic Least Squares Method, Weighted Least Squares Method, Logarithmic Least Absolute Values Method and Singular Value Decomposition~\cite{SAATY1984-comp, GASS2004-SVD, TAKEDA1987-evm, Blankmeyer1987ApproachesTC, brunelli2014introduction}. 
	The most used one is the Eigen Vector Method.  
	The EVM method computes the weight vector $w$ by solving the characteristic equation:
	\begin{equation}
		\label{eq:ahp:w}
		\left\{
		\begin{array}{l}
			{\bf A}\cdot w = \lambda_{max} \cdot w \\ 
			w^T {\bf 1} = 1
		\end{array}	\right.
	\end{equation} 
	where {\bf A} is the pairwise comparison matrix, $\lambda_{max}$ is the highest eigen value of {\bf A}, and 
	${\bf 1} = (1,\dots,1)^T$.
	The constraint $\sum_{i=1}^{m} w_i = 1$ is added to aid the solving process and avoid the infinitely many solutions. 
	Note that, if {\bf A} is positive then the highest eigenvalue is real (i.e. $\lambda_{max} \in \mathbb{R}$)~\cite{brunelli2014introduction}.
	Interestingly, Saaty~\cite{SAATY1977} mentioned that the perfect\footnote{When the components of {\bf A} are exactly obtained as ratio between weights.} normalized eigenvector $w$ satisfied the following relation:
	
	\begin{equation}
		\label{eq:ahp:relation}
		a_{ij} = \frac{w_i}{w_j},\, \forall i, j = 1 \ldots m.
	\end{equation} 
	
	In this case $\lambda_{max}=m$, otherwise $\lambda_{max} >  m$. The relation given by equation~(\ref{eq:ahp:relation}) explains why some researchers prefer the distance-based methods to \textit{directly} consider to minimize the distance between 
	$(a_{ij})_{m\times m}$ and $(w_{i}/w_{j})_{m\times m}$. 
	However, the resulting problem is non-linear and difficult to solve. 
	This is why we prefer, in this paper, to use the EVM method instead of distance-based methods.

\end{enumerate}

Obviously, the hierarchy may contain more levels of criteria, especially 
when the number of criteria exceeds nine. 
One key idea to handle this particular case consists to establish a hierarchy structure based on existing criteria dependencies. 
It is important to generate an optimized decision problem hierarchy, to reduce the number of pairwise decisions.
For more details, we refer the reader to~\cite{brunelli2014introduction}. 
Note that, in this paper, we will not cover this point.

\begin{example}
\label{exple:ahp}
Consider a multicriteria decision problem with three criteria, for which we build a pairwise comparison matrix {\bf A}$(a_{ij})_{3\times 3}$ that compares the relative importance of criteria according to the achievement of an overall goal. 
This matrix is given by: 
$$ {\bf A} =
\begin{pmatrix}
	1 & \frac{1}{2} & \frac{1}{4}\\
	2 & 1 & \frac{1}{2} \\
	4 & 2 & 1
\end{pmatrix}
$$

To find the criteria weight vector, one could use the EVM method, which consists to solve the equation: 
${\bf A} w = \lambda_{max} w$. The solution is given by the vector 
$w = (w_1, w_2, w_3) = (\frac{1}{7}, \frac{2}{7}, \frac{4}{7}) $ whose components are the weights of criteria.    
\end{example}

\newcommand{\kendall}{{\tt kendallW}}
\section{AHP-based learning approach for ranking patterns}
\label{sec-AHP-based-learning}

\newcommand{\learnWeights}{{\tt learnWeights}}

\newcommand{\learning}{{\tt computeCorrelation}}

\newcommand{\buildMatrix}{{\tt buildAHPMatrix}}

\newcommand{\SBH}{{\tt BaseHeuristic}}

\newcommand{\weight}{{\tt computeWeights}}
	\SetKw{return}{return}

Our objectif is to learn a ranking function aggregating a set of interestingness measures while approximating the user-specific preferences.

\subsection{Problem statement: learning pattern rankings}

We show in this section that learning a ranking function from some interestingness measures w.r.t. user-specific preferences can be translated into a multicriteria problem.
In our context, the criteria are interestingness measures.
%
 The task is to learn a function for ranking patterns from a sample of ordered patterns. Following previous research~\cite{DzyubaLNR14}, we use \textit{ordered feedback}, where a user is asked to provide a total order over a (small) number of patterns according to her subjective interestingness on them. 

Let $\triangleright$ be a binary preference relation between patterns.
A ranking function based on $\triangleright$, denoted by $r_{\triangleright}(.)$, accepts as input a subset of patterns ${\cal P} \subseteq \rels$ and returns as output a permutation $S$ $=$ $\langle P_{r_1}, P_{r_2}, \ldots,P_{r_n}\rangle$ 
 of $\mathcal{P}$, 
s.t. 
$\card{\mathcal{P}}=\card{S}$ and $\forall i < j, $ $P_{r_i} \triangleright P_{r_j}$.

The user ranking function to learn is denoted by $r_{\triangleright_u}$, which is based on the user preference relation $\triangleright_{u}$. 
Given a set of interestingness measures $\mathcal{M}=\{M_1, \ldots M_m\}$ and the corresponding ranking functions $r_{\triangleright_{M_1}}, \ldots, r_{\triangleright_{M_m}}$.
We denote by $rank_{M_i}(P_j)$ the rank of $P_j\in \mathcal{P}$ w.r.t. the interestingness measure $M_i$.
Here, $(P_k \triangleright_{M_i} P_l)$ means that $M_i(P_k)\geq M_i(P_l)$.

We propose to state the problem as a learning problem, where we have to find the weights $w_i$ of a linear aggregation function 
over the measures from a set of ranked patterns that maximizes the correlation of the aggregation with the user ranking function $r_{\triangleright_u}$:

\begin{equation}\label{eq:aggregation}
	\resizebox{0.3\hsize}{!}{$
		f(P)=\sum_{i=1..m} w_i \,.\,M_i(P)$}•
\end{equation}\\


Our main idea is to exploit the user feedback to perform pairwise comparisons between measures in order to estimate their relative importance on the basis of individual scores of measures. The results of these pairwise comparisons are then collected into a pairwise comparison matrix used by AHP to compute the weights $w_i$.

\paragraph{\bf Comparing a single  measure to a user ranking: }
\

Let $S= \langle P_{1}, \ldots, P_{n}\rangle$ be a user ranking.
Let $\mathcal{M}=\{M_1, \ldots, M_m \}$ be some interestingness pattern measures. 
To evaluate the overall ranking accuracy of a given measure compared to a user ranking, we use the Kendall's W concordance coefficient \cite{kendall1939}. 
{Given a measure $M_i$ and a user ranking $S$, we denote by $K_i(S)\in [0,1]$, the Kendall's W  between the ranking using $M_i$ and the user ranking $S$, where the value $1$ (resp., value $0$)
represents a perfect concordance (resp., no concordance): }



\begin{equation}\label{equ-kendall-w-formula}
\resizebox{0.5\hsize}{!}{$
K_i(S) = \frac{3\alpha}{({\abs{S}}^{3}-{\abs{S}})},\quad 
\alpha=\sum _{\ell=1}^{\abs{S}} (R_{\ell}-{\bar {R}})^{2}, $}
\end{equation}

\noindent
\resizebox{0.7\hsize}{!}{
where $R_{\ell} = rank_{M_i}(P_\ell) + rank_{u}(P_\ell)$ with $P_\ell\in S$, 
and $\bar{R} = \frac {1}{\abs{S}}\sum\limits_{\ell=1}^{\abs{S}} R_{\ell}$.
}

\paragraph{\bf Comparing two  measures according to a user ranking:}
\

Let $M_i$ and $M_j$ two measures to be compared according to a given user ranking $S$.
Computing the gap $\Delta_{i,j}(S)= K_{i}(S)-K_{j}(S)$ enables to know how much $M_i$ is closer to the user ranking $S$ than $M_j$: if $\Delta_{i,j}(S)\geq 0$, then $M_i$ is closer, else the converse. If a set of user rankings $\mathcal{S}=\{S_1,\ldots,S_n\}$ is considered, comparing two measures $M_i$ and $M_j$ on $\mathcal{S}$ comes to estimating the closest measure to $\mathcal{S}$ by averaging the gaps on all the user rankings of $\mathcal{S}$: \resizebox{0.3\hsize}{!}{$\Delta_{i,j}=\frac{1}{n} \sum\limits_{S_k\in\mathcal{S}} \Delta_{i,j}(S_k)$}.

\subsection{Learning weights process}

Function \ref{alg:AHPRank} implements the learning process that computes a weight vector $w$ (a weight for each measure) by exploiting the pairwise  comparisons of measures. 
It takes as inputs a vector of pairs comparisons $\Delta$ and the corresponding index $l$, a user ranking $S$ and a set of measures $\mathcal{M}$.
The learning step is done at line \ref{f:learn} where for each pair of measures, we compute the gap w.r.t the user ranking $S$.
$\learnWeights$ is an incremental function that can take into account the previous $l$ user rankings by updating $\Delta$.  
At line \ref{f:build},the AHP matrix is built by scaling the $\Delta_{i,j}$ comparisons to the permitted values (between ${1}/{9}$ and 9).  
Once done, the criteria weight vector $w$ is returned in line \ref{ahp:learnWeights} by solving the minimization problem described in line \ref{f:weight}, using the usual eigen vector method (EVM) \cite{brunelli2014introduction}.

\SetKwInOut{InOutput}{In Out}
\SetKwInOut{Input}{In}
\SetKwInOut{Output}{Out}

\begin{algorithm}

	\LinesNumbered

	\SetAlgoLined
 \floatname{algorithm}{Function}

	\Input{Set of measures $\mathcal{M}=\{M_1,\ldots, M_m\}$; 
		user ranking $S= \langle P_{1}, \ldots, P_{n}\rangle$;}
	\InOutput{ pairs comparisons and index parameter $\langle\Delta, l\rangle$;}
	\Output{Weight vector $w$;}

	
	
		
	\BlankLine
	\BlankLine
			${\bf A}[i,j] \gets 1,\,   \forall i,j \in \{1,\ldots,m\}$; 

	\BlankLine
	\BlankLine
		
			\ForEach{$M_i,M_j\in \mathcal{M}: i<j$}{\label{f:learn}
				$\Delta_{i,j}\gets \frac{l}{l+1} \Delta_{i,j}+ \frac{1}{l+1}(K_{i}(S)- K_{j}(S))$;\ \label{learn:kendall}
			}
			
				\BlankLine
				\BlankLine

		\ForEach{$\Delta_{i,j}\in \Delta$}{\label{f:build}
			\lIf{ $\Delta_{i,j} < 0 $}{
				
				\quad scale $\Delta_{i,j}$ to $(-9..-1);\quad$
				${\bf A} [j,i] \gets \abs{\Delta_{i,j}};\quad$
				${\bf A} [i,j] \gets 1/\abs{\Delta_{i,j}}$
			}
			\lElse{
				
				\quad scale $\Delta_{i,j}$ to $(1..9);\quad$ 
				${\bf A}[i,j] \gets \Delta_{i,j}; \quad$
				${\bf A}[j,i] \gets 1/\Delta_{i,j}$					
			}
		}	
		$w \gets solve\left(
		\begin{array}{l}
			{\bf A}\cdot w = \lambda_{max} \cdot w \\
			\text { subject to } \sum_{j=1}^{m} w_{j}=1, \quad w_{j}>0, \forall j=1 \ldots m .
\end{array}\right)$ \label{f:weight}
		
\Return{$w$;}\label{ahp:learnWeights}

	\caption{$\learnWeights(\langle\Delta, l\rangle, S, \mathcal{M})$\label{ahp:returnlearnweights}}\label{alg:AHPRank}
	
\end{algorithm}

Once the weight vector $w$ is learned, we can compute the score of a given pattern $P_i$ using the following weighted aggregation function:

\begin{equation}
	g_{w}(P_i)=\sum\limits_{M_i\in \mathcal{M}}w_i\ scale({M_i}(P)) 
	\label{eq:score}
\end{equation}  

where the $scale$ function is used to adjust interestingness values of the different measures to a $[0,1]$ scale.


\subsection{AHPRank algorithm}
\label{sec-AHP-based-learning}
Function \ref{alg:AHPRank} implements the key concept of our approach. It enables to learn a weight vector $w$ from a given user ranking $S$.  
Moreover, 
function \ref{alg:AHPRank} can easily be used in an iterative process and for a passive/active perspective
using the incremental computation of $\Delta$ in line $\ref{learn:kendall}$.
%
We propose \ahp{} (algorithm \ref{alg:ActiveLearning}) based on $\learnWeights$ function to learn a weighted aggregation function in a passive mode as well as in an active one.
Algorithm \ref{alg:AHPRank} takes as input a set of measures $\mathcal{M}$, a set of user-ranked patterns $\mathcal{S}$ and a triplet of parameter $\langle \mathcal{P},\theta,T\rangle$ corresponding  respectively to a collection of patterns and two integers to use in the active mode.
\ahp{} starts by initializing the vector of pairs comparisons to zero in line \ref{ahp:init}.
If \ahp{} is fed with a non empty $\mathcal{S}$, then it acts in a passive mode by iterating and calling $\learnWeights$ on the given user-ranked patterns $\mathcal{S}$ (lines \ref{ahp:startpassive}-\ref{ahp:endpassive}). 
Otherwise, the active mode is triggered by submitting $T$ queries to the user (lines \ref{ahp:startactive}-\ref{ahp:endactive}).
Here, the user is 
%
asked to rank a subset of patterns proposed by a query generator.
Our query generator is based on a heuristic that exploits the performance of the overall approach on the already learned rankings.
More precisely, the active learning mode is performed through the following main iterative steps:
\begin{enumerate}

\item As a first step, a subset of patterns is randomly sampled from the global set of patterns $\mathcal{P}$: At line \ref{ahp:sample}, ${\tt SamplePatterns}$ returns a sample $\mathcal{P'}$ of $\theta$ patterns.
\item Select a pair of patterns using ${\tt BasedHeuristic}$ function.
The proposed heuristic aims to select a pair of patterns of good quality to speed up the convergence of the learning process (function \ref{alg:sbh} detailed in section \ref{sec-heuristic}). 
\item Interacting with the user to get her preferences w.r.t to the patterns ranking query feedback (line \ref{ahp:feedback}).
\item Preference-based learning of the user ranking function (line \ref{ahp:dolearning}), by calling $\learnWeights$ function.
\end{enumerate}

It is important to stress that it is very usual that $\mathcal{P}$ is of huge size making step 2 very costly. Step 1 enables us to avoid this complexity by sampling a reasonable subset of patterns. Moreover, it also introduces a kind of diversity on the patterns proposed to the user through step 2.

To be effective, our learning method must decide when to stop e.g. maximal number of iterations, albeit being open to any custom stopping criterion. We take into account the case where the user can stop at any time she feels satisfied, and we simulate the user's stopping point using a fixed number $T$ of learning iterations.

\begin{algorithm}
	\LinesNumbered
	\SetAlgoLined
	 \setcounter{AlgoLine}{0}
	\Input{Set of measures $\mathcal{M}=\{M_1,\ldots, M_m\}$;\\
	Passive mode ($\mathcal{S} \neq \emptyset$): Set of user-ranked patterns $\mathcal{S}=\{S_1,\ldots, S_n\}$; \\
	Active mode ($\mathcal{S} = \emptyset$): Collection of patterns $\mathcal{P}$; Sample size $\theta$ ;\\ Number of iterations $T$; }
	
	
	\BlankLine

%

\lForEach{$\Delta_{i,j}\in\Delta$}{$\Delta_{i,j}\gets 0$}\label{ahp:init}

	\If{$\mathcal{S} \neq \emptyset$}{

\ForEach{$S_k\in \mathcal{S}$}{ \label{ahp:startpassive}

$w^t \gets {\tt LearnWeights}(\langle\Delta, k\rangle,S_k, \mathcal{M})$ \tcp*{Learn weights} \label{ahp:dopassivelearning}

}

	} \label{ahp:endpassive}
	\Else { \label{ahp:startactive}
		${w}^0 \gets \mathbf{1}$; \tcp*{Weight vector }  
		
\For{$t = 1, 2 \ldots T$}{	
$\mathcal{P'}\gets {\tt SamplePatterns}(\mathcal{P},\theta)$;\label{ahp:sample}

$Q_t \gets {\tt SensitivityBasedHeuristic}(\mathcal{P'}, {w}^{t-1}, \mathcal{M})$;\label{ahp:heuristic}

$S_{t} \gets {\tt AskRanking}(Q_{t})$ \tcp*{Ask user to rank the pair of patterns in $Q_t$} \label{ahp:feedback}
$w^t \gets {\tt LearnWeights}(\langle\Delta, t\rangle,S^*_t, \mathcal{M})$ \tcp*{Learn weights} \label{ahp:dolearning}
}
} \label{ahp:endactive}

\caption{{${\tt  AHPRank}(\mathcal{M}, \mathcal{S}, \langle\mathcal{P}, \theta, T\rangle)$}}\label{alg:ActiveLearning}
	
\end{algorithm}

\begin{algorithm}
	
	\setcounter{AlgoLine}{0}
	\LinesNumbered
	
	\SetAlgoLined
	\floatname{algorithm}{Function}
	
	\Input{Collection of patterns $\mathcal{P}$; Weight vector $w$; Set of measures $\mathcal{M}$; 
	}
	
	\BlankLine

$\mathcal{P'} \gets {\tt Sort}(\mathcal{P},w)$ \tcp*{sort $\mathcal{P}$ according to $g_w$}

$Pair \gets \emptyset$;

$\sigma_{min} \gets +\infty$;

\For{$i \in 1..\card{\mathcal{P'}}: P_i,P_{i+1}\in \mathcal{P'}$}{
	
	$score \gets \sigma(P_i,P_{i+1})$;\\
	\If{$score < \sigma_{min}$}{
		$\sigma_{min} \gets score$;\\
		$Pair\gets \{P_i,P_{i+1}\}$;\\
	}	
	
}	

\Return{$Pair$}; 	 \label{sbh:end}
	
	\caption{${\tt BasedHeuristic}(\mathcal{P}, w, \mathcal{M})$}\label{alg:sbh}
	
\end{algorithm}

\subsubsection{A Pattern Selection Heuristic for AHPRank}
\label{sec-heuristic}
Learning in active manner is not an easy task, since it is crucial that each learning step improves the learned function. Selecting the right patterns to present to the user is what allows the algorithm to improve the learning process step by step. However generating the right set of patterns is NP-hard \cite{DBLP:journals/jmlr/Ailon12}. 
We propose a heuristic (Function~\ref{alg:sbh}) based on sensitivity analysis for AHP models~\cite{10.1093/imaman/3.1.61}.
The rationale behind our heuristic is to select a single pair of patterns that are close in terms of overall score predicted by the learned function, relative to the sum of the measures gaps:

\begin{equation}
     \sigma(P_i,P_j)=\left\lvert \frac{g_w(P_i) - g_w(P_j)}{\sum_{l=1}^{m}( M_{l}(P_i) - M_{l}(P_j) )}\right\rvert ; i,j \in 1 \ldots k.
 \end{equation}\\
 
Intuitively, a successful aggregation should have the following property: if two patterns are close in the overall predicted score, they should be close also in their measures values. Thus, the algorithm will be enforced into learning the correct preferences of the patterns when having the lowest $\sigma$ value, since the selected pair is surely close in terms of $g_w$ values and very distant on the measures values.

\begin{example}
	Given four interestingness measures $\mathcal{M}=\{M_1,M_2,M_3,M_4\}$ and two patterns $\{P_1, P_2\}$.
	 Suppose that the measure values for both patterns are as follows: $P_1 =\langle 0.1,0.2,0.3,0.7 \rangle$, $P_2 =\langle 0.7,0.3,0.2,0.1 \rangle$
	and that the current learned weight vector is $w=\langle 0.25,0.25,0.25,0.25 \rangle$.
	At this stage, $g_w(P_1)=0.26$ and $g_w(P_2)=0.26$.
	Here $P_1$ and $P_2$ are equivalent w.r.t. the current state of $g_w$.
	However, the values of the four measures are different for the two patterns. 
	This pair of patterns is an interesting candidate ($\sigma(P_1,P_2)=0$), where eliciting a preference on such a pair will make the learning closer to the user ranking.
\end{example}

\subsection{Complexity Analysis} \label{sec:complexity}
\begin{proposition}[\bf Time complexity] 
Given a bounded number of interestingness measures, Algorithm~\ref{alg:ActiveLearning} runs in $O(n\ k)$  in passive learning and in $O(n)$ in active learning. 
\end{proposition}

\emph{\textbf{proof}}
Let $\mathcal{S}=\{S_1,\ldots,S_n\}$ be a set of $n$ user rankings,
 $k$ the size of the largest ranking $S_t$ and $m$ a bounded  number of measures in $\mathcal{M}$.
Once the pairwise comparison matrix ${\bf A}$ is built, we could compute the preference vector of weights $w$ using several mathematical techniques.
We have used the eigen-vector based method (i.e., EVM) \cite{brunelli2014introduction}, for which the worst case time complexity 
 is about $O(m^3)$ \cite{Eigenproblem99}.
The EVM approach is of constant time complexity $O(1)$ since that $m$ is constant.
Complexity of computing the matrix at line \ref{learn:kendall} of the $\learnWeights$ function is ${O}(m\ n\ k)$ since that Kendall's W is in ${O}(k)$ (see Definition (\ref{equ-kendall-w-formula})). 
The complexity of lines  $\ref{f:learn}$, $\ref{f:build}$ and $\ref{f:weight}$ in $\learnWeights$ are respectively ${O}(n\ m^2 + m\ n\ k)$, ${O}(m^2)$ and ${O}(m^3)$. 
Notice that in AHP, it is demonstrated in \cite{saaty2003magic} that the  number  of  criteria is recommended to be  no  more than seven $\pm 2$. 
That is, with $m \leq 9$, we get an asymptotic quadratic complexity of ${O}(n\ k)$ when \ahp{} is in passive mode. In other terms, the approach costs ${O}(k)$ for each ranking in the given $\mathcal{S}$.
Now, when \ahp{} is in active mode where queries are pair of patterns ($k=2$), the complexity is linear on the number of queries submitted to the user $O(n)$.

It is important to stress that in practice, we have in general a set $\mathcal{S}$ reduced to a large element in the passive learning mode ( $n=1$ and a large $k$), and
to a set of $n$ queries of pair of patterns (a large $n$ and a $k=2$) in the active mode.
This makes \ahp{} running in a linear time (linear on $k$ in passive and on $n$ in active) in most of the cases. 
This low complexity makes \ahp{} a fast approach to learn a user ranking function, which is supported by the experimental evaluation in section~\ref{sec:xp}.

\section{Running Example}
\newcolumntype{Y}{>{\centering\arraybackslash}X}

To illustrate our approach, we consider an initial set of patterns $\mathcal{P}=\{P_1,\ldots,P_{10}\}$, 
five interestingness measures $\mathcal{M}=\{M_1,\ldots, M_5\}$ and a user-ranking $r_{\triangleright_u}$.
Columns of Table \ref{tab:patterns} show the user ranking for $\mathcal{P}$, the rankings given by $M_i$ on $\mathcal{P}$ and the \ahp{} results.  
It is important to stress that no measure perfectly matches the user ranking. 

\begin{table}[h]
	\centering
	\begin{tabular} {|c|c||cc|cc|cc|cc|cc||cc|} 
		\hline
		$user$ & $S$ &    \multicolumn{2}{c|}{$M_1$ -- rank} &    \multicolumn{2}{c|}{$M_2$ -- rank} &    \multicolumn{2}{c|}{$M_3$ -- rank} &    \multicolumn{2}{c|}{$M_4$ -- rank}&   \multicolumn{2}{c||}{$M_5$ -- rank} &  \multicolumn{2}{c|}{$g_w$ -- rank} \\ 
		\hline
		 \textbf{1}    & $P_7$        &0.95 & \textbf{1}  & 0.48 & 8  & 0.79 & 2  & 0.30 & 9  & 0.80 & \textbf{1} &0.72  &\textbf{1} \\ 

		 \textbf{2}   &$P_3$       & 0.75 & 3  & 0.72 & 1  & 0.78 & 3  & 0.70 & \textbf{2}  & 0.61 & \textbf{2} & 0.68 &\textbf{2}\\ 

		 \textbf{3}   &$P_6$        & 0.80 & 2  & 0.49 & 7  & 0.50 & 9  & 0.65 & 4  & 0.60 & \textbf{3}  & 0.61 &\textbf{3}\\ 
		
		 \textbf{4}    & $P_1$       & 0.47 & 9  & 0.47 & 9  & 0.76 & 5  & 0.56 & 6  & 0.59 & \textbf{4}  & 0.54 & \textbf{4}\\ 
		
		 \textbf{5}    & $P_8$       & 0.56 & 6  & 0.65 & 4  & 0.63 & 8  & 0.69 & 3  & 0.40 & \textbf{5}  & 0.53 &\textbf{5} \\ 

		 \textbf{6}    & $P_{10}$       & 0.57 & 5  & 0.50 & \textbf{6}  & 0.80 & 1  & 0.4  & 8  & 0.02 & 10  & 0.34&8 \\

		 \textbf{7}    & $P_5$        & 0.62 & 4  & 0.62 & 5  & 0.66 & 6  & 0.57 & 5  & 0.27 & \textbf{7}  &0.48 &\textbf{7} \\ 
		
		 \textbf{8}    & $P_2$        & 0.48 & \textbf{8}  & 0.66 & 3  & 0.65 & 7  & 0.1  & 10 & 0.05 & 9  &0.33 &9 \\ 		
		
		 \textbf{9}    & $P_4$       & 0.50 & 7  & 0.68 & 2  & 0.77 & 4  & 0.50 & 7  & 0.35 & 6  & 0.50&6 \\ 
				
		 \textbf{10}    & $P_9$       & 0.02 & \textbf{10} & 0.1  & \textbf{10} & 0.05 & \textbf{10} & 0.8  & 1  & 0.25 & 8  &0.18 &\textbf{10} \\ 
		
		\hline
	\end{tabular}
	\caption{Running example with $10$ patterns and $5$ measures.\label{tab:patterns}} 
\end{table}

The data in Table \ref{tab:patterns} is used by our algorithm to build the AHP Matrix and to learn a weight vector $w$ over $\mathcal{M}$. 
For the sake of simplicity, let us call \ahp{} in the passive mode with 
$\mathcal{S}=\{S_a=\langle P_3,P_1,P_5,P_2,P_4\rangle \}$.
Here, the user prefers $P_3$ $\triangleright_u$ $P_1$ $\triangleright_u$ $P_5$ $\triangleright_u$ $P_2$ $\triangleright_u$  $P_4$.
\ahp{} learns weights by calling $\learnWeights$ on $S_a$ and  by computing the correlation between the user rankings of $S_a$ and the ones of the measures $M_i$.
The matrix provided in Equation~(\ref{eqn:delta-mat}) shows the $\Delta$ values of each measure pairs.
We recall that for $\vert \mathcal{S} \vert > 1$, $\Delta_{i,j}$ is an averaged value of $\Delta_{i,j}(S_k)$ where $S_k\in \mathcal{S}$.
Since we only have $S_a$, $\Delta_{1,2} = \Delta_{1,2}(S_a)$.
Here, $\Delta_{1,2}(S_a)$ represents the gap between the two measure rankings $M_1$ and $M_2$  w.r.t the user ranking $S_a$.
For having $\Delta_{1,2}(S_a)$, we need to compute Kendall's W  $K_{1}(S_a)$ (resp., $K_{2}(S_a)$) between the ranking of $M_1$ (resp. $M_2$) compared to the user ranking $S_a$: $\Delta_{1,2}(S_a)= (K_{1}(S_a)-K_{2}(S_a))= (0.25 - 0.25) = 0$.
After scaling the values of $\Delta$ to $(-9..-1)$ if negatives, and to $(1..9)$ otherwise. We can see in Equation~(\ref{eqn:delta-mat}) that, for instance, measure $M_5$ is better than $M_4$ with a degree of $6$. The same intensity is observed between $M_5$ and $M_3$.
We can also observe that $M_3$ and $M_4$ are indifferent.

\renewcommand{\kbldelim}{(}
\renewcommand{\kbrdelim}{)}
{\footnotesize
	\begin{equation}
		\label{eqn:delta-mat}
		\Delta = \kbordermatrix{
			& (M1) & (M2) & (M3) & (M4) & (M5) \\
			(M1) &  & 0 &	0.42 &	0.42 &	-0.43 \\
			(M2)&& &	0.20 &	0.28&-0.25\\
			(M3)&&   &	&	-0.08&-0.55\\
			(M4)& &	&	&	&-0.60\\
			(M5)&&	 &	 &	 &
		} \overset{Scaling}{\longrightarrow} 
		\kbordermatrix{
			& (M1) & (M2) & (M3) & (M4) & (M5) \\
			(M1) &  &1 &	4 &	4 &	-4 \\
			(M2)&& &	2 &	3&-3\\
			(M3)&&   &	&	1&-6\\
			(M4)& &	&	&	&-6\\
			(M5)&&	 &	 &	 &
		} 
	\end{equation}
}

Afterwards, \ahp{} computes the AHP matrix $A$ through $\learnWeights$ function and the scaled $\Delta$ matrix (i.e., the average correlation gap) as an input:

\renewcommand{\kbldelim}{(}
\renewcommand{\kbrdelim}{)}
{\small
	\begin{equation}
		\label{eqn:AHP-mat}
		{\bf A}
		= \kbordermatrix{
			& (M1) & (M2) & (M3) & (M4) & (M5) \\
			(M1)&1&	1 &	4&	4&1/4\\
			(M2)&1&	1&	2&	3&1/3\\
			(M3)&1/4&  1/2&	1&	1&1/6\\
			(M4)& 1/4&	1/3&	1&	1&1/6\\
			(M5)&4&	 3&	 6&	 6&1
		}
	\end{equation}
}

At the end, \ahp{} computes the weighting vector $w$ 
by solving the minimization problem.
Thus, we find the learned weight vector $w = (w_{M_1}, w_{M_2}, w_{M_3}, w_{M_4}, w_{M_5})$ $=$ $(0.24, 0.24, 0.065, 0.065, 0.39)$. 
This vector reflects the importance of each measure with respect to the achievement of the goal (user ranking function $g_w$).

Now, we can use the AHP interestingness measure $g_w$ of formula \ref{eq:score} to rank all the patterns ${\cal P}$ provided in Table \ref{tab:patterns}.
The overall ranking accuracy of \ahp{} on such example is of $91\%$ where it is able to rank accurately the 5 first patterns, the seventh and the tenth ones.   

%
%
%
%
%
%
%
%
%
%
%

\newcommand{\ones}{x^{-1}& (1)\xspace}
\newcommand{\zeros}{x^{-1}& (0)\xspace}
\newcommand{\unbound}{x^{-1}& (*)\xspace}

\newtheorem{defi}{Definition}
\newtheorem{prop}{Proposition}
\newtheorem{corolarry}{Corollary}

\let\comment=\relax
\def\trackingLevel{2}
\newcommand{\comment}[3]{\ifnumcomp{\trackingLevel}{=}{2}{{\color{#1}[\bf\footnotesize{{#2: \textit{#3}}}]}}{}}


\newcommand{\zoo}{{$\tt Zoo$}\xspace}
\newcommand{\Hepatitis}{{$\tt Hepatitis$}\xspace}

\newcommand{\vote}{{$\tt Vote$}\xspace}
\newcommand{\anneal}{{$\tt Anneal$}\xspace}
\newcommand{\chess}{{$\tt Chess$}\xspace}
\newcommand{\mushroom}{{$\tt Mushroom$}\xspace}
\newcommand{\connect}{{$\tt Connect$}\xspace}
\newcommand{\tone}{{$\tt T10$}\xspace}

\newcommand{\tfour}{{$\tt T40$}\xspace}

\newcommand{\pumsb}{{$\tt Pumsb$}\xspace}
\newcommand{\retail}{{$\tt Retail$}\xspace}

\newcommand{\zooI}[2]{{$\tt Zoo\_#1\_#2$}}
\newcommand{\voteI}[2]{{$\tt Vote\_#1\_#2$}}
\newcommand{\annealI}[2]{{$\tt Anneal\_#1\_#2$}}
\newcommand{\chessI}[2]{{$\tt Chess\_#1\_#2$}}
\newcommand{\mushroomI}[2]{{$\tt Mushroom\_#1\_#2$}}
\newcommand{\connectI}[2]{{$\tt Connect\_#1\_#2$}}
\newcommand{\toneI}[2]{{$\tt T10\_#1\_#2$}}
\newcommand{\tfourI}[2]{{$\tt T40\_#1\_#2$}}
\newcommand{\pumsbI}[2]{{$\tt Pumsb\_#1\_#2$}}
\newcommand{\retailI}[2]{{$\tt Retail\_#1\_#2$}}

\newcommand{\rand}{{$\tt rand$}\xspace}
\newcommand{\lex}{{$\tt lex$}\xspace}

\newcommand{\cputimee}{{\tt t}}
\newcommand{\recc}[1]{${\tt R@{#1}}$}
\newcommand{\rec}[1]{${\tt R@{#1\%}}$}

\section{Experiments}
\label{sec:xp}
In this section, we experimentally evaluate our fast learning framework  \ahp{} for ranking patterns. 
First, we present our case study on association rules mining and different oracles simulating user-specific rankings.
Then, we present the studied research questions, the experimental protocol and the obtained results.

\newcommand{\Rule}[2]{{#1}\rightarrow {#2}}

\subsection{Mining associations rules (ARs)} 
\label{sec:case_study}
We evaluate our generic approach on a concrete pattern mining task, the association rule mining, one of the most important and well
studied task in data mining, first introduced in \cite{DBLP:conf/sigmod/AgrawalIS93}. 

An association rule  is an implication of the 
form $\Rule{X}{Y}$, where $X$ and $Y$ are itemsets such that 
$X\cap Y=\emptyset$ and $Y \neq \emptyset$. 
$X$ represents the body of the rule and $Y$ represents its head.
The \emph{frequency} of an itemset $X$ in a dataset, denoted by $freq(X)$, 
is the number of transaction of the dataset containing $X$.
The frequency of a rule $\Rule{X}{Y}$ is the frequency of the itemset $X\cup Y$, 
that is,  $freq(\Rule{X}{Y})=freq(X\cup Y)$.

Several interestingness measures of ARs have been introduced like support, confidence, interest factor, correlation and entropy. Tan et al. \cite{tan2004selecting} made an interesting study on the usefulness of the existing measures related to the application type. In particular, they identified seven independent groups (see Table \ref{tab:plain}) of consistent measures having similar properties.

\definecolor{Gray}{gray}{0.85}
\newcolumntype{Y}{>{\centering\arraybackslash}X}
\newcolumntype{Z}{>{\raggedleft\arraybackslash}X}
\newcolumntype{a}{>{\columncolor{Gray}}c}

\begin{table}[h]

	\begin{tabularx}{\textwidth}{ccX}
		\hline
		 Groups &$\qquad$ & Measures                                                                  \\ \hline
		1  & &\textbf{Yules Q}, Yules Y, Odds Ratio                                               \\ \hline
		2    &   & \textbf{Cosine}, Jaccard                                                           \\ \hline
		3    &   & \textbf{Laplace}, Support                                                          \\ \hline
		4    &   & \textbf{$\phi$coefficient}, CollectiveStrength, piatetskyShapiro's \\ \hline
		5    &   & \textbf{Goodman Kruskal's}, Gini Index                                             \\ \hline
		6    &   & \textbf{Interest factor},added value, Klosgen K                                    \\ \hline
		7    &   & \textbf{certainty factor}, Mutual Information, Cohen's $\kappa$                           \\ \hline
	\end{tabularx}

	\caption{Independent Subjective Measure Groups}
	\label{tab:plain}
\end{table}

For our evaluation, we pick one measure per group to have $7$ measures representing an independent set of measures (see measures in bold in  table \ref{tab:plain}).

\subsection{User feedback emulators}
It is difficult to evaluate an interactive approach since users are scarce, for this reason we emulate the user feedback using three different objective target ranking functions: 
\newcommand{\randemu}{{\sc Rand-Emu}}
\newcommand{\lexemu}{{\sc Lex-Emu}}
\newcommand{\chiemu}{{\sc Chi-Emu}}

\begin{itemize}
	\item \randemu: The user-specific ranking is equivalent to a random weighted aggregation function. For that, we generate for each measure $M_i$ a random weight $w_i\in [0,1]$ such that $\sum\limits_{i\in 1..m} w_i=1$.\\
	\item \lexemu: The user-specific ranking follows a lexicographic order on the measures. That is, given a lexicographic order $lex$ on $\mathcal{M}$: $$lex(\mathcal{M})=\langle l_1, \ldots, l_m \rangle, s.t., \bigcup\limits_{i\in 1..m} l_i = \mathcal{M}$$ 
	Here, given two patterns $(P_1, P_2)$, $P_1$ is prefered to $P_2$ iff $(l_i(P_1)>l_i(P_2))$ or $(l_i(P_1)=l_i(P_2) \wedge l_{i+1}(P_1)>l_{i+1}(P_2))$ (for $i = 1$ to $m-1$).\\
	
	\item \chiemu:  $\chi^2$ as a statistical measure is a good candidate to emulate the user feedback \cite{RINGUEST1986379}. $\chi^2$ is a rather complex function to approximate with non-trivial correlations, it is also a quality measure suggested in \cite{dzyuba2017learning}. Here, we use $\chi^2$ as the target user-specific ranking function over ARs. Given some association rule $\Rule{X}{Y}$, the $\chi^2$ value is as follows:
	
	\begin{equation}
		\chi^2 (\Rule{X}{Y}) = \frac{ (freq(\Rule{X}{Y})- \frac{freq(X) freq (Y)}{N})^2}{\frac{freq (X) freq (Y)}{N}}.
	\end{equation}
	with $N$ the number of transactions of the dataset. 
 
\end{itemize}

\newcommand{\RQ}[1]{\textbf{RQ#1}}

\subsection{Research questions}

Our evaluation aims to answer the following four research questions:

\begin{itemize}
	\item \RQ{1}: \emph{Given a sample of ranked patterns, is it possible to infer the user-specific preferences over all the patterns? If yes, how much of data are required as input? How long does the learning process last?}
	\item \RQ{2}: \emph{How does the proposed new learning based on AHP compare with the baseline one using SVM when they are used to automatically learn user-specific ranking function?}
	\item \RQ{3}: \emph{How effective is \ahp{} from an active learning perspective? Is the sensitivity heuristic a wise choice to use in query selection?} 
	\item \RQ{4}: \emph{How effective is \ahp{} from an interactive data mining perspective?} 
\end{itemize}

\subsection{Experimental protocol}

\subsubsection{Implementation settings}
We have implemented in Java our \ahp{} approach with its two modes: the passive mode denoted by \ahpp{} and the active mode denoted by \ahpa{}.
The code is publicly available at ${\tt github.com/lirmm/AHPRank}$.
We compared our approach to the state of the art, the SVM based approach  
\svm{} \cite{dzyuba2017learning}.
We denote by \svmp{} the passive version and by \svma{} the active one following \cite{dzyuba2013interactive}.
All experiments were conducted on an Intel core $i7$, $2.4Ghz$ with $16Gb$ of RAM using an overall timeout of one hour.

\subsubsection*{Metrics}
We consider three metrics to evaluate the performance of our approach:
\begin{itemize}
\item[(i)] the Spearman's rank correlation coefficient $\rho$ (sum of squared differences between learned $rank_L$ and target $rank_T$ pattern ranks over $n$ patterns)
in order to evaluate the overall ranking accuracy:

\begin{equation}
	{\displaystyle \rho= 1-{\frac {6\sum\limits_{i\in 1..n} (rank_{L}(P_i)-rank_{T}(P_i))^{2}}{n(n^{2}-1)}},}
\end{equation}

\item[(ii)] the Recall metric \recc{k} in order to evaluate the effectiveness of our approach in identifying the top $k$ most interesting patterns:

\begin{equation}
	{\tt R}@k = \frac{\abs{\{rank_{L}(P_i) \leq k:  i \in 1..n \wedge rank_{T}(P_i) \leq k\}}}{k}
\end{equation}
\item[(iii)] CPU time needed to finish the whole learning process and the waiting time between two queries for the active mode (in seconds).

\end{itemize}

\subsection{Benchmark Datasets}

We selected several real-sized datasets from the FIMI repository.\footnote{fimi.uantwerpen.be/data/} 
These datasets have various characteristics representing different application domains. 
Table ~\ref{tab:datasets} reports for each dataset the number of transactions \nbTrans, the number
of items \nbItems, 
its application domain,
and the number of  valid rules  (\nbRules) corresponding to the initial rules minded
using a standard association rules algorithm and without any knowledge about the user.
The datasets are presented by increasing order of \nbRules.

\begin{table}[]
	
	\centering
	
	\begin{tabularx}{\textwidth}{|Y|Y|Y|Y|c||Y|} \hline
		
		{Dataset} & \multicolumn{1}{c|}{\nbTrans} & \multicolumn{1}{c|}{\nbItems}   & {Density(\%)}& {Type of Data} & \multicolumn{1}{c|}{\nbRules}   \\
		\hline
		\hline
		{\Hepatitis}  & 137 & 68 &50.00&{Disease} & 0.5 M\\
		{\connect}  & 67,557  & 129 &33.33& {Game steps} &1 M\\
		{\mushroom}  & 8,124  & 119 &18.75& {Species of mushrooms} & 1.5 M\\
		{\tfour}  & 100,000  & 1,000 &4.20&{Synthetic dataset} &2 M\\
		{\retail}  & 88,162 & 16,470 &0.06& {Retail market basket data} & 2,5 M\\ \hline
		\multicolumn{6}{r}{ \tfour:{$\tt T40I10D100K$}}
	\end{tabularx}

	\caption{Dataset Characteristics. \label{tab:datasets}} 

\end{table}

\newcommand{\vbm}{${\tt VBM}$}

\newcommand{\timeout}[1]{${\tt TO} $}

\newcommand{\cputime}{${\tt Time}\,$}
\newcommand{\latencytime}{${\tt TQ}\,$}

\subsection{Passive learning results}

\begin{table}[h]
	\centering
	\begin{tabularx}{\textwidth}{|l||Z||Z|Z|Z|Z|Z|Z|Z|} 
		\hline 
		\multirow{2}{*}{\textbf{Datasets}}	&	\multicolumn{8}{|c|}{\textbf{measures}} \\ \cline{2-9}
		&\multicolumn{1}{c|}{(1)}  & \multicolumn{1}{c|}{(2)} 
		& \multicolumn{1}{c|}{(3)} & \multicolumn{1}{c|}{(4)}  
		& \multicolumn{1}{c|}{(5)} &  \multicolumn{1}{c|}{(6)} 
		& \multicolumn{1}{c|}{(7)}& \vbm \\  \hline \hline
		\multicolumn{9}{|c|}{\randemu}\\ \hline	
		{\Hepatitis} &0.78
		&0.92
		&0.29
		&\textbf{0.97}
		&0.79
		&0.85
		&0.45
		&\textbf{0.97}   
		\\ 
		
		{\connect} 
		& 0.68
		& 0.62
		& \textbf{0.95}
		& 0.70
		& 0.60
		& 0.45
		& 0.57   
		&\textbf{0.95}		
		\\ 
		
		{\mushroom} 
		&{0.84}
		&\textbf{0.97}
		&0.28	
		&0.83
		&0.55 
		&0.73	
		&0.36
		&\textbf{0.97}
		\\

		{\tfour}&0.71
		&\textbf{0.99}
		&0
		&0.76
		&\textbf{0.99}
		&\textbf{0.99}
		&0  
		&\textbf{0.99}	
		\\ 
		
		{\retail} &0
		&\textbf{0.98}
		&0.43
		&0.71
		&0.84
		&\textbf{0.98}
		&0.51 
		&\textbf{0.98}		
		\\ \hline

		\multicolumn{9}{|c|}{\lexemu}\\ \hline
		{\Hepatitis} &0.79
		&0.64
		& 0
		&0.68
		&\textbf{0.92}
		&0.63
		&0   
		&\textbf{0.92 }
		\\ 
		
		{\connect} 
		
		&0.21
		&0.47 
		&\textbf{0.76}
		&0.26 
		& 0 
		&0.23
		&0.50 
		&\textbf{0.76}		
		\\ 
		
		{\mushroom} &0.21 
		&\textbf{0.82}
		&0.38	
		&0.77
		&0.78
		&0.37
		&0
		&\textbf{0.82}
		\\

		{\tfour}&0.76
		&0.98
		&0
		&0.77
		&\textbf{0.99}
		&0.97
		&0  
		&\textbf{0.99}	
		\\ 
		
		{\retail} &0.10
		&0.78
		&0.54
		&0.49
		&\textbf{0.84}
		&0.80
		&0.58 
		&\textbf{0.84}		
		\\ \hline 

		\multicolumn{9}{|c|}{\chiemu}\\ \hline
		{\Hepatitis} &0.20
		&	0.92
		&	0.36
		&	\textbf{0.96}
		&0.73
		&0.85
		&0.43  
		&\textbf{0.96}  
		\\ 

		{\connect} 
		&0.09
		&0.15
		& 0 
		&0.02
		& \textbf{0.29}
		& 0.01
		& 0
		&\textbf{0.29}		
		\\ 
		
		{\mushroom} &0.51 
		&0.64
		& 0	
		&0.46	
		&0.28 
		&\textbf{0.88}	
		& 0 
		&\textbf{0.88}
		\\

		{\tfour}&0.71
		&0.98
		&0.17
		&0.64
		&0.98
		&\textbf{0.99}
		&0.29	
		&\textbf{0.99}		
		\\ 
		
		{\retail} &0
		&\textbf{0.94}
		&0.53
		&0.60
		&0.83
		&0.57
		&0.61 
		&\textbf{0.94} \\  \hline

		\multicolumn{9}{|r|}{(1): \yules $\qquad$(2): \cosine $\qquad$ (3): \lap$\qquad$ (4): \lev }\\
		\multicolumn{9}{|r|}{  (5): \lamb $\qquad$ (6): \IF$\qquad$ (7): \cert  }\\ 
		\hline
	\end{tabularx}
	\caption{Correlation results with user ranking for using the emulators. }
	\label{tab:xp1}
\end{table}
In this section, we address the two first research questions (\RQ{1} and \RQ{2}). 
To that end, we perform a 5-folds cross-validation on \nbRules{} for each dataset.
In each fold, we randomly select $20\%$ of \nbRules{} to form the training data and use the remaining $80\%$ ARs for evaluation (testing data). We have chosen this way of using k-folds cross-validation in order to see how effective are the approaches in learning from relatively small training sets.

\medskip
\noindent
\textbf{A) Analysing the different user feedback emulators.} 
Table \ref{tab:xp1} reports the rank correlation $\rho$ between the user-specific ranking functions (\randemu, \lexemu{} and \chiemu) and the seven interestingness measures. 
We also report \vbm{}, the \emph{Virtual Best Measure}, which returns the best rank correlation $\rho$ provided by one of the seven measures. 
We observe from table \ref{tab:xp1} that given a measure, the results are chaotic.
For instance, \lamb{} measure is highly correlated to \chiemu{} on \tfour{} dataset ($\rho=98\%$) but it is weakly correlated on \connect{} ($\rho=29\%$) dataset. 
The same observation can be made with \randemu{} and \lexemu{} functions.
However, the theoretical construction of \vbm{} is of a high accuracy (mean of $91\%$), which brings us to believe that a weighted aggregation of the selected measures can lead to a good trade-off.

\begin{table}
	\centering	
	
	\begin{tabularx}{\textwidth}{|l||Z|Z|Z|Z||Z|Z|Z|Z|} 
		\hline
		\multicolumn{9}{|c|}{\textbf{\randemu}} \\ \hline
		
		&  \multicolumn{4}{c||}{\svmp} & \multicolumn{4}{c|}{ \ahpp}\\ \hline
		
		&\multicolumn{1}{c|}{$\rho$}&\multicolumn{1}{c|}{\rec{10}}& \multicolumn{1}{c|}{\rec{1}} & \multicolumn{1}{c||}{t(s)}
		&\multicolumn{1}{c|}{$\rho$}&\multicolumn{1}{c|}{\rec{10}}& \multicolumn{1}{c|}{\rec{1}} & \multicolumn{1}{c|}{t(s)}		
		\\ \hline
		{\Hepatitis } 
		&0.94&0.74&0.77&851&0.93&0.79&0.71&9 
		\\ 
		{\connect } 
		&  {-}&-&-& \tt TO &0.95 &0.72&0.63&26 
		\\ 
		{\mushroom }  
		& {-}&-&-& \tt TO&0.89&0.69&0.66&32
		\\ 
		{\tfour  }
		&{-}&-&-& \tt TO&0.99&0.93&0.58 & 63
		\\ 
		{\retail }
		&{-}&-&-& \tt TO&0.93&0.95&0.96&66
		\\ \hline \hline
		{{$mean$}} & - & - & -& -& 0.94&0.82&0.71& 39\\ \hline 		\hline
		
		\multicolumn{9}{|c|}{\textbf{\lexemu}} \\ \hline

		&  \multicolumn{4}{c||}{\svmp} & \multicolumn{4}{c||}{ \ahpp}\\ \hline
		
		&\multicolumn{1}{c|}{$\rho$}&\multicolumn{1}{c|}{\rec{10}}& \multicolumn{1}{c|}{\rec{1}} & \multicolumn{1}{c||}{t(s)}
		&\multicolumn{1}{c|}{$\rho$}&\multicolumn{1}{c|}{\rec{10}}& \multicolumn{1}{c|}{\rec{1}} & \multicolumn{1}{c|}{t(s)}		
		\\  \hline
		{\Hepatitis  } 
		&0.99&0.99&0.99&902&0.92&0.86&0.81& 13\\ 
		{\connect  } 
		&  {-}&-&-& \tt TO &0.60 &0.46&0.50&21\\ 
		{\mushroom  }  
		& {-}&-&-& \tt TO&0.86&0.56&0.65&32\\ 
		{\tfour  }
		&{-}&-&-& \tt TO&0.99&0.93&0.46 & 71\\ 
		{\retail  }
		&{-}&-&-& \tt TO&0.78&0.68&0.65&68\\ \hline \hline
		{{$mean$}} & -&-&-& - & 0.83 &0.70&0.61 & 41\\ \hline 		\hline
		
		\multicolumn{9}{|c|}{\textbf{\chiemu}} \\ \hline
		
		&  \multicolumn{4}{c||}{\svmp} & \multicolumn{4}{c|}{ \ahpp}\\ \hline
		
		&\multicolumn{1}{c|}{$\rho$}&\multicolumn{1}{c|}{\rec{10}}& \multicolumn{1}{c|}{\rec{1}} & \multicolumn{1}{c||}{t(s)}
		&\multicolumn{1}{c|}{$\rho$}&\multicolumn{1}{c|}{\rec{10}}& \multicolumn{1}{c|}{\rec{1}} & \multicolumn{1}{c|}{t(s)}		
		\\ \hline
		{\Hepatitis  } 
		&0.99& 0.97& 0.92 & 979 &{0.94}&{0.90}& 0.77 & 10\\ 
		{\connect  } 
		& {-}&-&-& \tt TO & 0.15 &{0.28}&{0.67}& 17\\ 
		{\mushroom  }  
		& {-}&-&-& \tt TO & 0.98&0.91&0.93 & 28\\ 
		{\tfour  }
		& {-}&-&-& \tt TO&0.99&0.99&0.96& 84\\ 
		{\retail  }
		& {-}&-&-& \tt TO& 0.91&0.96&0.95 & 50\\ \hline \hline
		{{$mean$}} &-&-&-& - & 0.79 &0.81&0.86 & 38\\ \hline 		\hline
		
	\end{tabularx}
	\caption{5-folds cross-validation results (passive learning).}
	\label{tab:xp2}
\end{table}

\medskip
\noindent
\textbf{B) Comparing \ahpp{} with \svmp{}.} 
Table \ref{tab:xp2} is dedicated to the k-folds cross-validation results of \svmp{} and \ahpp{}
using averaged rank correlation $\rho$, recall values (\rec{10} and \rec{1}) and CPU time in seconds averaged over the folds. 
Under a time contrast of one hour, \svmp{} is able to deal with a training set not exceeding $100K$ rules, and thus we are able to compare it with our approach only on \Hepatitis{}.
In terms of ranking accuracy, we observe that \svmp{} outperforms \ahpp{} for all user feedback emulators.
However, \ahpp{} remains competitive with an acceptable accuracy. 
Same observations can be made in terms of recall at the $10\%$ and $1\%$ top of the ranking (\rec{10} and \rec{1}).
\ahpp{} is able to reach a high correlation with the user ranking functions on most of the datasets.
However, we can also observe a weak correlation with \chiemu{} on \connect{}, where a high correlation is observed on that dataset with \randemu{} and \lexemu{}. This stems from the fact that \randemu{} and \lexemu{} are linear functions expressed with the given $7$ interestingness measures, which explains the high accuracy.
However, \chiemu{} is a complex function and the statistics extracted from \connect{} on the $7$ measures are not sufficient to learn such a function.   
However, the high accuracy of \svmp{} is at the expense of the running time.
For a training set of $100K$ rules, \svmp{} needs more than $15$ minutes to learn.
Exceeding $100K$ rules, \svmp{} needs more than one hour,
where \ahpp{} is able to deal with $7.5M$ of rules in a time not exceeding $4$ minutes.\\

In what follows, the observations and the conclusions drawn from \chiemu{} remain true for \randemu{} and \lexemu.
For the sake of simplicity, we only report the results on the complex function \chiemu{}.

\begin{figure}[H]
	\centering
	\includegraphics[width=\textwidth]{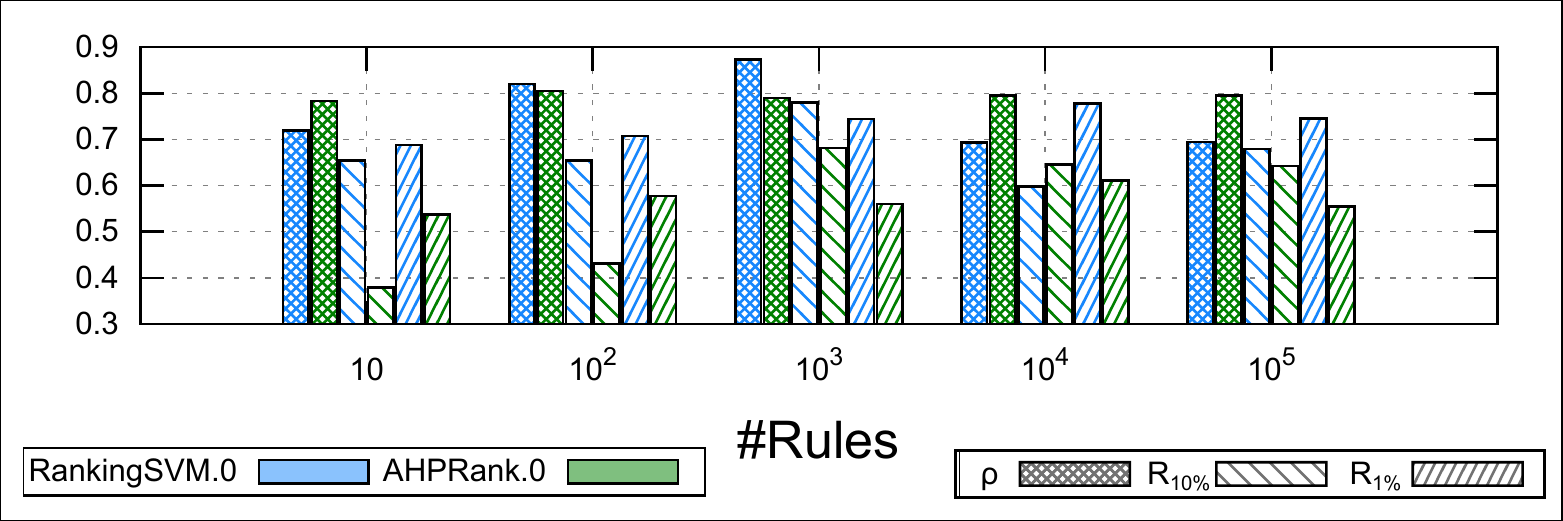} %
	
	\caption{ Learning accuracy comparison between \svmp{} and \ahpp{}  learning \chiemu{} on different training data size.}%
	\label{fig:passive1}%
\end{figure}

\begin{figure}[H]
	\centering
	
	\includegraphics[width=\textwidth]{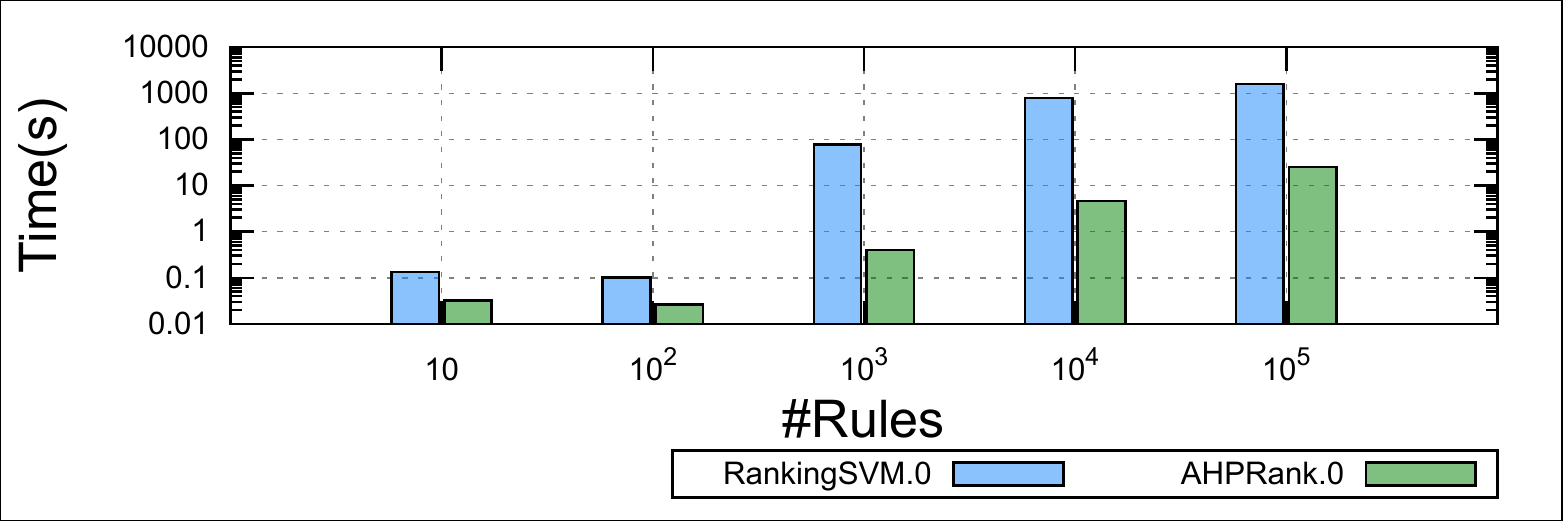} %
	\caption{ CPU time comparison between \svmp{} and  \ahpp{}  learning using \chiemu{} emulator on different training data sizes averaged overall datasets.}%
	\label{fig:passive2}%
\end{figure}

\medskip
\noindent
\textbf{C) Impact of varying the size of the training data on the learning.} 
It is important to stress that the proposed \ahp{} aims to ensure a fast learning process to reveal a pattern ranking function, while offering good accuracy.
To strengthen our observation regarding the performance scalability of \ahpp{} compared to \svmp{}, 
we report in figures \ref{fig:passive1}  and  \ref{fig:passive2} a performance comparison by varying the size of the training data on learning \chiemu{} function.
Similar findings have been made on \randemu{} and \lexemu{} functions.
For each dataset, we select randomly $nb$ rules and we call the two approaches ($\ nb\in\{10, 100, 1K, 10K, 100K\}$).
The results are averaged on ten runs.

In terms of Spearman correlation $\rho$, figure \ref{fig:passive1} shows  a discrepancy of $5\%$ between the two approaches when the training data is not exceeding $1K$.
However, the gap becomes particularly important (exceeding $10\%$) and in favor of \ahpp{} when the training data contains $10K$ and $100K$ rules.
In terms of recall at $10\%$ and $1\%$, \svmp{} outperforms \ahpp{} with a gap of $23\%$ and $15\%$ at \rec{10} and \rec{1} when the training data contains only $10$ rules. The gap get tighter when the training data is increasing in size at \rec{10} (less than $5\%$) and remains more or less the same for \rec{1}.

In terms of CPU time,  figure \ref{fig:passive2} shows that \svmp{} needs less than a second to deal with a training data not exceeding $100$ rules, more than one minute for $1K$ rules, $13min$ for $10K$ rules, and more than $26min$ for $100K$ rules. For comparison, \ahpp{} is able to deal with a training data of a size ranged between $100$ and $100K$ in a time ranged between $0.03$ and $24.86$ seconds
for an overall ranking accuracy close to the one reported by \svm.

\subsection{Active learning results}

In this section, we attempt to answer the two research questions  (\RQ{2} and \RQ{3}) where  we compare the active learning versions of \svm{} and \ahp{}.
For that end, we consider a simple process, where we ask the user some ranking queries on pairs of patterns (e.g.  \emph{do you prefer $P_i$ to $P_j$?}). 

\medskip
\noindent
\textbf{A) Evaluating the effectiveness of our sensitivity-based heuristic.} 
Our first experiment aims to evaluate the usefulness of our Sensitivity-Based Generator (SBG)  (see section ~\ref{sec-AHP-based-learning}).
For that, we compare SBG to a Random Generator (RG), where the latter selects randomly a pair of patterns from \nbRules{} to submit to the both learners \svma{} and \ahpa{}. 
For our experiment and with a human-in-the-loop, we set the number of queries (i.e., number of iterations $T$ of algorithm \ref{alg:ActiveLearning}) to a reasonable number not exceeding $20$ queries.
We repeat $10$ times the experiment and take the average result in order to avoid the bad-luck effect with RG and the sampling step of our SBG.
After few tests, we set the size of the sample $\mathcal{X}$ picked at line ~\ref{ahp:sample} of algorithm ~\ref{alg:ActiveLearning} to $\theta=10^3$.
A sample size that provide a good trade-off between time selection and the accuracy of the selected pair. 

\begin{figure}
	\centering
	\includegraphics[scale=0.6]{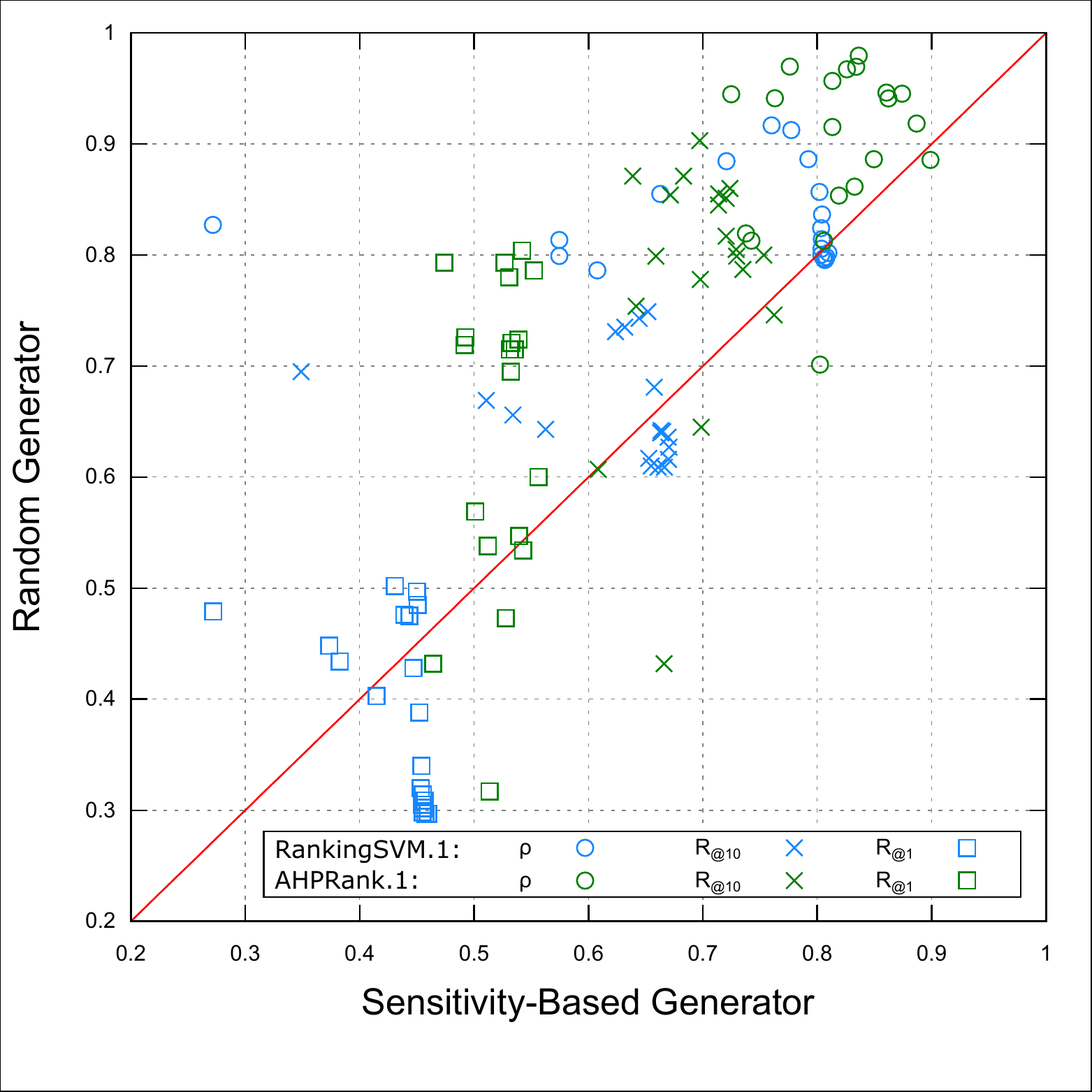} %
	
	\caption{ Random vs Sensitivity-Based Patterns Generator.}%
	\label{fig:active}%
\end{figure}

Figure \ref{fig:active} reports a comparison between \svma{} and \ahpa{} in a scatter plot of $20$ iterations.
We report the Spearman correlation $\rho$  and the recall at $10\%$ and $1\%$ of each iteration.
We observe from the scatter plot that the use of SBG outperforms RG either with our approach or with \svma{}.

\newcommand{\ci}{${\tt CI}$}


In order to support  our observation on the use of SBG, we carried out a statistical test using the {\em Wilcoxon Signed-Rank Test}, 
Here, the one-tailed alternative hypothesis is used with the following null hypothesis (\emph{RG is more efficient than SBG}):
$H_0:$ The accuracy using RG $\geq$ The accuracy  using SBG.
That is, the alternative hypothesis $H_1$ states that our SBG outperforms RG.
With this statistical test, we are able to conclude that the use of SBG is more efficient than RG ($H_1$ accepted).
Table \ref{tab:wilcoxon} reports the $p$-$value$, the $z$-$score$ and the confidence interval (\ci) of each test.
According to the $p$-$value$ column, and except the case of (\svma,\rec{10}), we have strong evidence to reject the null hypothesis.
The confidence interval column shows clearly that the use of SBG is better than RG.
Consequently, we will use in what follows the SBG heuristic in \svma{} and \ahpa.

\begin{table}[h]
	\centering
	
	\begin{tabularx}{\textwidth}{|Y|Y|YYY|} 
\hline
\textbf{Approaches} & \textbf{Metrics} & $p$-$value$ & $z$-$score$ & \ci	\\ 
\hline		
\multirow{3}{*}{\svma} & $\rho$ & $.00226$& $-2.8373$&  $99\%$\\ 
 & \rec{10} & $.12302$ & $-1.1573$ & $70\%$ \\  
  & \rec{1} & $.06301$ & $-1.5306$& $90\%$ \\ \hline
  \multirow{3}{*}{\ahpa} & $\rho$ & $.00034$& $-3.3973$& $99.9\%$ \\ 
  & \rec{10} & $.00205$ & $-2.8746$& $99\%$\\ 
  & \rec{1} &$.00139$ & $-2.9866$ & $99\%$\\ \hline
	\end{tabularx}

	\caption{ Wilcoxon Signed-Rank Test (RG vs SBG).} 		\label{tab:wilcoxon}
\end{table}

\begin{table}
	\centering
	
	\begin{tabularx}{\textwidth}{|r|r|XXXXX|} 
		\hline
		\multicolumn{2}{|c|}{}  & (a)& (b)& (c)& (d)& (e)      \\ 
		\hline \hline
		\multicolumn{7}{|c|}{ \textbf{10 Queries} } \\ \hline
		
		\multirow{4}{*}{\rotatebox[origin=c]{0}{(1)} }  
		
		& $\rho$  & 0.73& -0.04& 0.69& {0.99}& 0.97\\ 
		& \rec{10} & 0.61& \textbf{0.28}& 0.38&{0.92}& {0.91} \\ 
		& \rec{1}  & 0.44& 0.49& {0.36}& 0.45& 0.67\\ 
		& \cputime     & {0}        & {0}       & {0}         & {0}         & {0}              \\ 
		\hline
		
		\multirow{4}{*}{\rotatebox[origin=c]{0}{(2)} }  
		
		& $\rho$  &\textbf{0.78}& \textbf{0.10}& \textbf{0.93}& {0.99}& \textbf{0.99}\\ 
		& \rec{10} & \textbf{0.66}& 0.36& \textbf{0.80}&{0.97}& {0.91} \\ 
		& \rec{1}  & \textbf{0.51}& \textbf{0.50}& \textbf{0.71}& \textbf{0.81}& \textbf{0.93}\\ 
		\hline

		\hline

		\multicolumn{7}{|c|}{ \textbf{50 Queries}} \\ \hline
		
		\multirow{4}{*}{\rotatebox[origin=c]{0}{(1)} }  
		
		& $\rho$  & 0.94& \textbf{0.20}& 0.71& {0.99}& {0.98}\\ 
		& \rec{10} & 0.77& 0.10& 0.39& 0.92& {0.91} \\ 
		& \rec{1}  & 0.57& 0& {0.36}& 0.45& 0.67\\ 
		& \cputime   & \textbf{0} & \textbf{0} & \textbf{0} & 2 & 3\\
		\hline
		
		\multirow{4}{*}{\rotatebox[origin=c]{0}{(2)} }  
		& $\rho$  & \textbf{0.95}& 0.10 & \textbf{0.93}& {0.99}& \textbf{0.99}\\ 
		& \rec{10} & \textbf{0.83}& \textbf{0.40}& \textbf{0.72}& \textbf{0.94}& {0.91} \\ 
		& \rec{1}  & \textbf{0.66}& \textbf{0.45}& {0.51}& \textbf{0.64}& \textbf{0.93}\\ 
		\hline

		\multicolumn{7}{|c|}{ \textbf{100 Queries}} \\ \hline
		
		\multirow{4}{*}{\rotatebox[origin=c]{0}{(1)} }  
		
		& $\rho$  & 0.94& \textbf{0.29}& 0.72& {0.99}& 0.98\\ 
		& \rec{10} & 0.75& 0.20& 0.41& 0.92& {0.91}\\ 
		& \rec{1}  & 0.53 & 0 & {0.36} & 0.45& 0.67\\ 
		& \cputime   & 0 & 2 & 3& 3& 10\\		\hline
		
		\multirow{4}{*}{\rotatebox[origin=c]{0}{(2)} }  
		& $\rho$  & \textbf{0.98}& 0.10& \textbf{0.81}& {0.99}& \textbf{0.99}\\ 
		& \rec{10} & \textbf{0.89}& \textbf{0.35}& \textbf{0.49}& \textbf{0.95}&{0.91}\\ 
		& \rec{1}  & \textbf{0.78} & \textbf{0.42}& \textbf{0.37} &\textbf{0.74}& \textbf{0.93}\\ 
		\hline
		\hline
		
		\multicolumn{7}{|r|}{(1): \svma $\qquad$ (2): \ahpa}\\ 
		
				\multicolumn{7}{|r|}{(a): \Hepatitis$\quad$ (b): \connect$\quad$ (c): \mushroom$\quad$ (d): \tfour$\quad$ (e):\retail}\\ \hline
		
	\end{tabularx}
	\caption{Qualitative evaluation of \svma{} vs \ahpa. \label{tab:active}}
\end{table}

\medskip
\noindent
\textbf{B) Comparing \ahpa{} with \svma{}.} 
Table~\ref{tab:active} shows a comparison between \svma{} and \ahpa{} on the $5$ datasets.
We report $\rho$, \rec{10} and \rec{1} within stopping criterion $T$ at $10$, $50$ and $100$ queries.
We also report the averaged latency time between two queries \cputime{} for \svma{}, knowing that \cputime never exceeds the latency bound of $0.1$ seconds. The findings are the same for \randemu{}, \lexemu{} and \chiemu{} functions. 

The main observation that we can draw from table \ref{tab:active} is that \ahpa{} outperforms \svma{}.
Let us take a close look at \Hepatitis{} results. 
For this dataset, \ahpa{} is able to rank $500K$ patterns with an accuracy of $78\%$ using only $10$ queries ($73\%$ observed with \svma).
The accuracy increases to $95\%$ when learning on $50$ queries for \ahpa{}  and $94\%$ for \svma{}. \ahpa{} reaches  $98\%$ of accuracy with $100$ queries while \svma{} remains stable at $94\%$.
In terms of recall values, \ahpa{} is able to discover the most relevant patterns on the first $50K$ and $5K$ patterns (out of $500K$) 
with an accuracy of, respectively, $78\%$  and $66\%$ over $10$ queries ($61\%$ and $44\%$ for \svma).
Let us now take a close look at \retail{} dataset.
We observe that \ahpa{} reaches a high accuracy over $10$ queries and such accuracy remains stable over the following $90$ queries.
The \svma{} results are less impressive on \retail{} and particularly on the recall metric.
But the main difference is on the waiting time \cputime , where \svma{} can keep the user waiting  more than $10$ seconds between two queries.
Same observation can be made on the other datasets where \svma{} can be hampered by overall waiting time even with queries of size $2$,
where \ahpa{} shows an instantaneous behavior (less than $0.1$ seconds between two queries).
This represents a limitation in the use of \svma{} especially when the learning is integrated to an interactive data mining  process, where a reasonable latency time for a human user is around few seconds \cite{Lallemand12}.

\newcommand{\err}{${\tt Err}$}
\subsection{Interactive learning results}
In this section, we address the last research questions \RQ{4}. 
In addition to the active learning results, which are part of the interactive perspective with a human-in-the-loop, we conduct two more experiments by focusing on the hazard characteristics linked to the presence of a human in the learning process.

Our first experiment aims to evaluate the robustness of our approach in front of a human likely to make mistakes.
To simulate such situations involving incorrect user feedback, we pick randomly (with a percentage ratio \err{}) a set of queries and we swap the user preference.
That is, a user that prefers a pattern $P_i$ to $P_j$ will make a mistake by preferring $P_j$ to $P_i$ with a probability \err{}.

\begin{figure}[h]
	\centering
	\includegraphics[width=\textwidth]{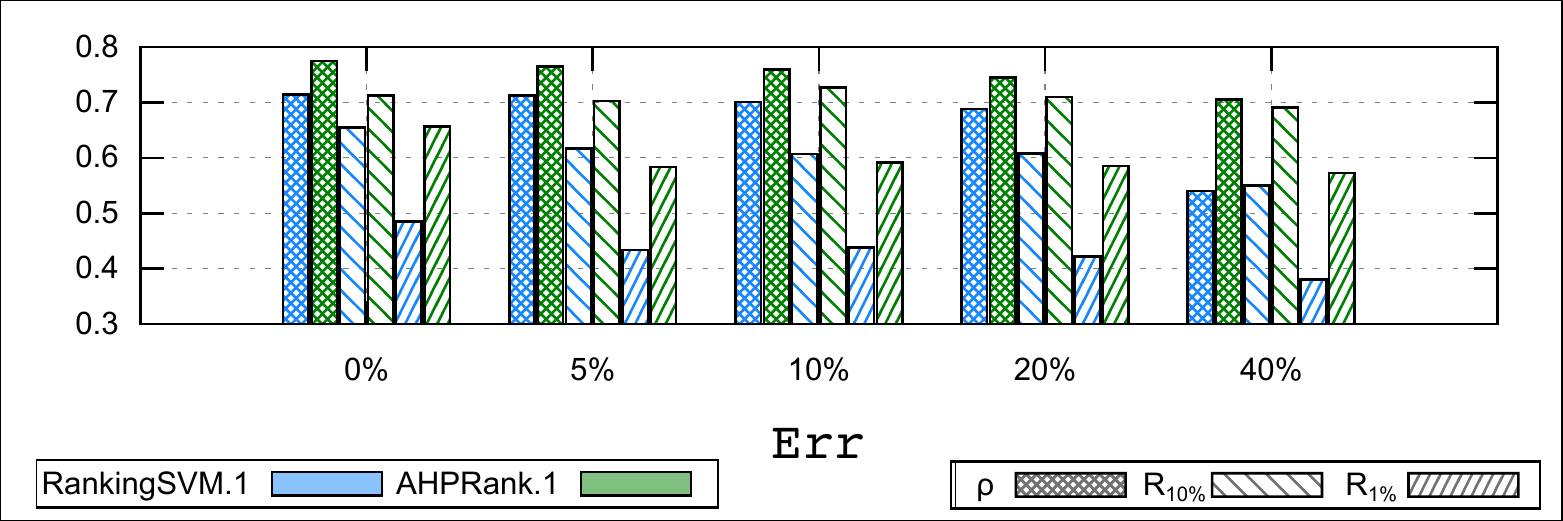} %
	
	\caption{ Learning under mistakes ($20$ queries).}%
	\label{fig:active-err}%
\end{figure}

Figure \ref{fig:active-err} reports a comparison between \svma{} and \ahpa{} under mistakes present with a probability \err{} varying from $0\%$ to $40\%$.
The reported results are averaged on $10$ runs and on the whole set of datasets on learning \chiemu{} and by submitting $20$ queries to the user.
We observe that \ahpa{} is quite stable and robust even with $40\%$ of mistakes ($8$ mistakes out of $20$).
The overall correlation between the learned function and the user one $\rho$ is at $77\%$ without mistakes and becomes $70\%$ under \err$=40\%$.
However, \svma{} is stable till \err$=20\%$ and then, the accuracy is highly impacted where $\rho$ decreases from $71\%$ to $54\%$.
In terms of recall, we observe a decline not exceeding $2\%$ at \rec{10} and $8\%$ at \rec{1} under \ahpa{}.
Using \svma{}, the decline is exceeding $10\%$ at \rec{10} and \rec{1}. 
In terms of CPU time and between two queries, we observe a waiting time ranged between $6$ and $12$ seconds under \svma{}, where under our approach is never over $0.02$ seconds.

\newcommand{\lexrand}{{\sc Lex2Rand}}

\begin{figure}[h]
	\centering
	\includegraphics[width=\textwidth]{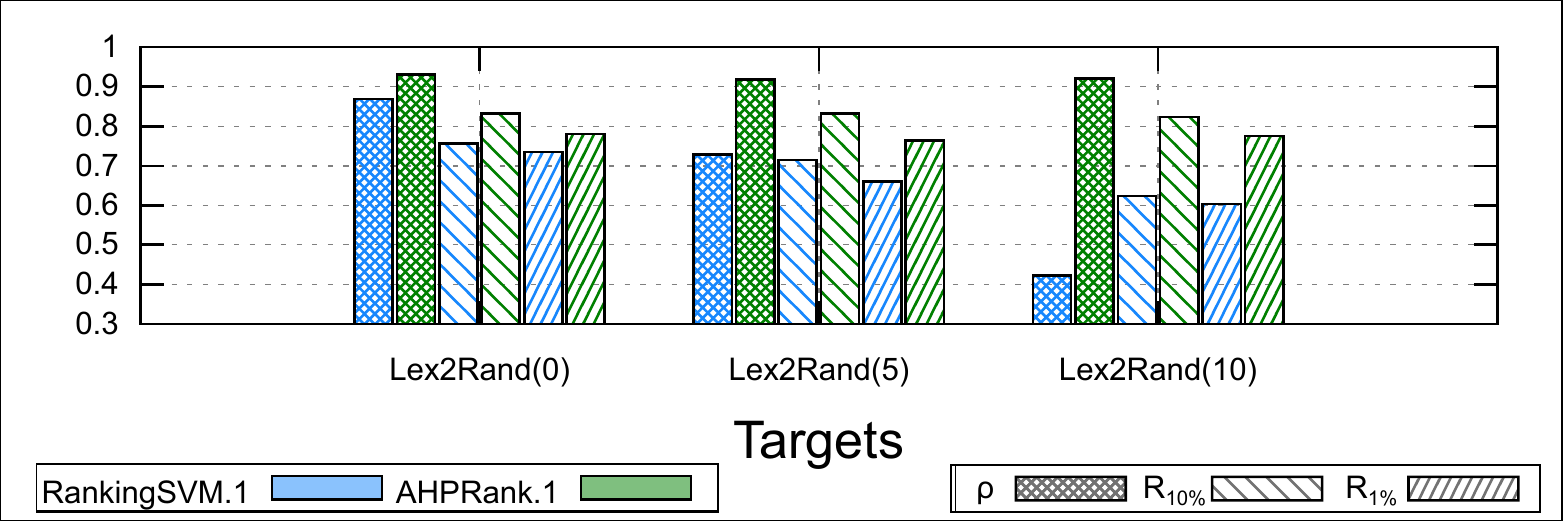} %
	
	\caption{ Learning under swaps ($20$ queries)}%
	\label{fig:active-swap}%
\end{figure}

Our second experiment aims to evaluate the robustness of our approach in front of an undecided user that starts with a set of preferences $A$ in mind to finish with preferences $B$. Here, the initial user preferences $A$ are biased along the presented patterns till being the user more comfortable with preferences $B$.
To simulate such situations where the user target function is changing during the learning process, we conduct an experiment where we start with \lexemu{} as a target to learn and after $x$ queries, we swap to \randemu{} where the learning process takes $(20-x)$ more queries (we coined it \lexrand{} target). 
Note that we choose a swap from \lexemu{} to \randemu{} because of the two functions are linear. 
For a total number of $20$ queries, we compare \svma{} and \ahpa{} on \lexrand{}($x$) target ($x\in\{0,5,10\}$).
Note that with $x=0$, \lexrand{(0)} is equivalent to \randemu{}.
Figure \ref{fig:active-swap} reports the averaged results of $10$ runs on all datasets over $20$ queries.
The main observation that we can draw is that \ahpa{} is stable in front of a changing linear function to learn even with a changeable mind in the middle of the learning ($10$ out of $20$ queries).
However, a target alteration during the learning process can drastically impacts the accuracy of \svma{} (decline of $44\%$ in terms of $\rho$). 
In terms of CPU time, our approach is $50$ times faster than \svma{}.

To sum up, our main finding is that \ahp{} represents a good trade-off between ranking accuracy and running time. 
\ahp{} is able to reach a good overall ranking accuracy within a time being linear on the size of the training data and thus, not exceeding few seconds, which makes it suitable for industrial exploitation. 

\section{Conclusion}
We presented a generic and efficient framework for learning pattern ranking functions using a multicriteria decision making method based on AHP. 
The proposed \ahp{} algorithm can be applied in both passive and active learning modes. It requires the user to rank subsets of patterns according to her preferences. 
Our approach exploits the learned weights to aggregate all the measures into a single ranking function, using a weighted linear formulation. 
The resulting aggregation function was experimentally and statistically proven to be as close as possible to the user ranking. 
We applied this framework to the case study of Association Rules Mining, for which experiments showed that \ahp{} algorithm  is able to learn the ranking function efficiently compared with the results of state-of-the-art learning approaches. 

\bibliographystyle{unsrt}  

\bibliography{reference.bib}

\end{document}